\pdfoutput=1
\documentclass[11pt]{article}

\usepackage{acl}
\usepackage{times}
\usepackage{latexsym}
\usepackage{makecell}
\usepackage{comment}
\usepackage{todonotes}
\usepackage{color}
\usepackage{tikz}
\usetikzlibrary{shapes,decorations,arrows,calc,arrows.meta,fit,positioning}

\tikzset{
    -Latex,auto,node distance =1 cm and 1 cm,semithick,
    state/.style ={ellipse, draw, minimum width = 0.7 cm},
    point/.style = {circle, draw, inner sep=0.04cm,fill,node contents={}},
    bidirected/.style={Latex-Latex,dashed},
    el/.style = {inner sep=2pt, align=left, sloped}
}

\usepackage[T1]{fontenc}
\usepackage[utf8]{inputenc}

\usepackage{microtype}
\usepackage{amsfonts}
\usepackage{amsmath}
\usepackage{amssymb}
\usepackage{breqn}
\usepackage{graphicx}
\usepackage{tikz}
\usepackage{xspace}
\usepackage{xcolor}
\usepackage{booktabs}
\usepackage{pdfpages}
\usepackage{comment}

\usepackage[shortlabels]{enumitem}
\usepackage{arydshln}
\usepackage{bm}
\usepackage{multirow}
\usepackage{algorithm}
\usepackage[noend]{algpseudocode}

\usepackage{amsfonts}
\usepackage{algpseudocode}

\makeatletter
\algnewcommand{\LineComment}[1]{\Statex \hskip\ALG@thistlm \(\triangleright\) #1}
\makeatother

\usepackage{microtype}

\title{Hierarchical Neural Data Synthesis for Semantic Parsing}

\author{
        Wei Yang,~~ 
        Peng Xu,~~
        Yanshuai Cao\\
  Borealis AI  \\
\texttt{\small \{wei.yang, peng.z.xu, yanshuai.cao\}@borealisai.com} 
  }
\begin{document}

\maketitle
\begin{abstract}
  Semantic parsing datasets are expensive to collect. 
%   Moreover, unlike many supervised learning problems where labels are scarce, the questions pertinent to a given domain might not be readily available, especially in cross-domain semantic parsing. 
  Moreover, even the questions pertinent to a given domain, which are the input of a semantic parsing system,  might not be readily available, especially in cross-domain semantic parsing. 
  This makes data augmentation even more challenging. Existing methods to synthesize new data use hand-crafted or induced rules, requiring substantial engineering effort and linguistic expertise to achieve good coverage and precision, which limits the scalability. In this work, we propose a purely neural approach of data augmentation for semantic parsing that completely removes the need for grammar engineering while achieving higher semantic parsing accuracy. Furthermore, our method can synthesize in the zero-shot setting, where only a new domain schema is available without any input-output examples of the new domain. On the Spider cross-domain text-to-SQL semantic parsing benchmark, we achieve the state-of-the-art performance on the development set ($77.2\%$ accuracy) using our zero-shot augmentation.
\end{abstract}

\section{Introduction}
Data augmentation improves generalization by injecting additional diversity into the training data without expensive manual curation. In computer vision, it is relatively simple to augment data by various image transformations \cite{lecun1998gradient,krizhevsky2012imagenet,shorten2019survey}. However, in other modalities like text, there are often no simple transformation rules to add invariance or equivariance while keeping the data natural. 

For NLP tasks requiring only coarse semantics, e.g. sentiment or topic classification, heuristics such as dropping words or synonyms replacement \cite{wei-zou-2019-eda} can be used. But on problems like semantic parsing, a small change in the input could alter its meaning and hence the output. Some prior approaches leverage ad-hoc heuristics or synchronous grammars \cite{andreas-2020-good,wei2019eda,wang-etal-2015-building,jia2016data}. The formers are too noisy while the latter can be cumbersome to implement. Some recent methods use only SQL grammar instead of synchronous grammar, and fill the questions with a learned neural net \cite{zhong-etal-2020-grounded,wang-etal-2021-learning-synthesize}, accepting the potential mistakes from the learned SQL-to-text model. These methods are noisier than synchronous grammar but can be easier to implement and achieve better coverage. 

This work proposes a pure neural data augmentation approach for cross-domain semantic parsing that has a better trade-off between coverage, precision, and ease of implementation. Instead of sampling SQL first and then generating questions, we produce questions hierarchically by learning to sample the entities given the domain schema, then generate questions using a fine-tuned text-to-text model conditioned on the entities. Our method then uses a modified self-training setup to achieve higher cross-domain semantic parsing accuracy. Furthermore, similar to \cite{zhong-etal-2020-grounded}, our method can be applied to new domain where only the schema or domain ontology is available, without any samples of question-SQL pairs. This ability for zero-shot augmentation is useful in bootstrapping semantic parsers on new domains. 
Empirically, with transductive augmentation, we achieve the state-of-the-art Exact Match (EM) accuracy of $77.2\%$ on the Spider development set. When transductive augmentation cannot be applied, our result of $71.8\%$ on the test set is competitive with the current state-of-the-art.

\begin{figure}[ht]
    \centering 
    \includegraphics[width=0.45\textwidth]{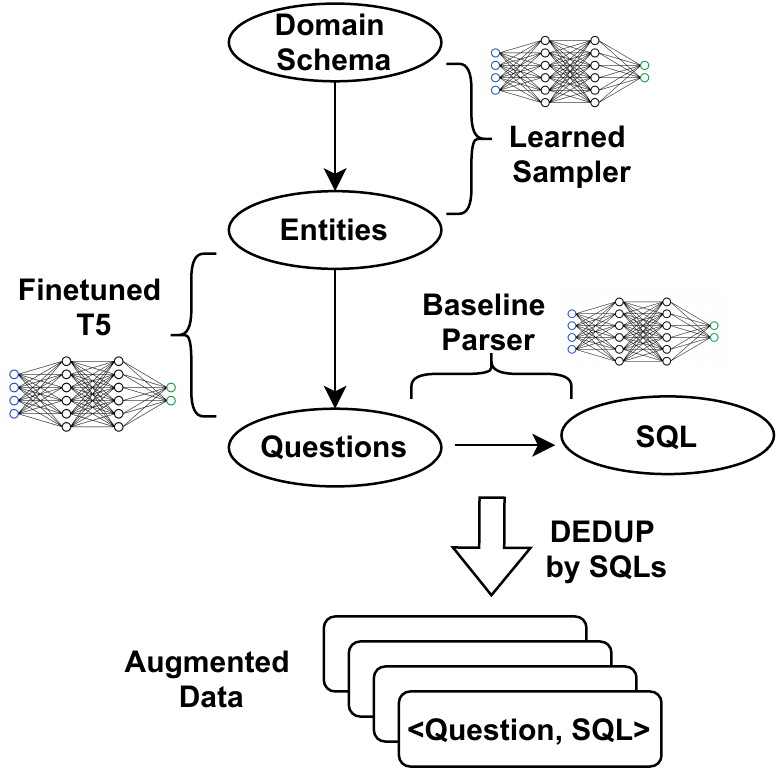}
    \caption{Our hierarchical neural data synthesis pipeline}
    \label{fig:framework}
\end{figure}

\section{Background: Trade-off in Data Augmentation for Semantic Parsing}

Semantic parsing requires complete and precise mapping between natural language utterances and their meaning representations. To ensure the augmented data is accurate, methods often rely on rules or grammars, which are either hand-crafted or induced from existing training data \cite{wang-etal-2015-building,andreas-2020-good,zhong-etal-2020-grounded,wang-etal-2021-learning-synthesize}. However, there is a three-way trade-off among the ease-of-implementation of such grammar systems, coverage of the augmentation, and the precision of logical forms. 

Both \citet{andreas-2020-good} and \citet{wei2019eda} use very simple heuristic rules to alter or recombine training examples. This class of methods is very easy to implement but has poor coverage and precision due to the limitation of the simple heuristics. While \citet{andreas-2020-good} demonstrates some empirical gains, its performance boost is much smaller compared to more sophisticated methods as shown in \citet{wang-etal-2021-learning-synthesize}.

\citet{wang-etal-2015-building} handcrafted a synchronous grammar that can produce utterance and meaning representation pairs and have human annotators rewrite the grammar-generated utterances to natural ones. If carefully designed, the synthesized dataset can have high coverage of semantic variations while achieving high precision. However, designing synchronous grammar requires linguistic expertise and may not be accessible to many practitioners, while the rewrite step requires further manual efforts. Hence, although \citet{wang-etal-2015-building} claims to build a semantic parser overnight, in practice, the effort required for the whole pipeline, including designing the synchronous grammar, is much more substantial.  
\citet{jia2016data} adopts a different trade-off by inducing the synchronous grammar from data instead of handcrafting. They achieve high precision with relatively lighter implementation effort but suffer in coverage. 

Alternatively, by using a grammar of the meaning representation only instead of asynchronous grammar, it is possible to simplify implementation further. This is the idea behind learning to perform SQL-to-text synthesis from \citet{zhong-etal-2020-grounded,wang-etal-2021-learning-synthesize}. Both approaches use a SQL grammar to sample queries first, then use a SQL-to-text model learned on the training dataset to translate sampled SQL queries to natural questions. \citet{zhong-etal-2020-grounded} uses a handcrafted SQL templates while \citet{wang-etal-2021-learning-synthesize} learns the distribution of queries. This class of methods further lowers the implementation challenge while accepting some noise in the augmentation due to errors in the learned SQL-to-text translation. 

Empirically, \citet{andreas-2020-good} previously demonstrated slight improvement over \citet{jia2016data} on GeoQuery, while \citet{wang-etal-2021-learning-synthesize} showed it is significantly superior to \citet{andreas-2020-good} on the same dataset. Therefore, we will focus our effort in comparing to \citet{wang-etal-2021-learning-synthesize} and \citet{zhong-etal-2020-grounded}, in the more challenging case of cross-domain semantic parsing \cite{yu2018spider}.
% Our approach completely removes the need for grammar while achieving an even better result than the previous ones. 

\section{Hierarchical Neural Synthesis Pipeline}
To give an overview of our method, first note that learning-to-synthesis \cite{zhong-etal-2020-grounded,wang-etal-2021-learning-synthesize} is already accepting noisy labels with the learned SQL-to-text mapping, then why not synthesize in the opposite direction by sampling questions first? 

To produce pertinent and diverse questions, we learn a cross-domain neural model that accepts a schema and samples a sequence of entities to appear in a question. Then we fine-tune a large pre-trained text-to-text model like T5 \cite{raffel2020exploring} to map from entity sequences to questions. Finally, we use a baseline semantic parser learned on the training set to annotate the generated question. Crucially, we use the baseline parser to filter beam-search results of the entity-to-question generation, retaining only those questions that have actual semantic variations and can be recognized by the baseline parser. 

The resulting pipeline does not require any additional grammar than what might already be in the baseline parser, hence easy to implement. The overall pipeline achieves great diversity and coverage because we can use large pre-trained models compared to sampling SQL queries first, which is limited to learning SQL distribution from a much smaller set. Finally, the synthesis is still low noise thanks to the beam filtering using the learned parser. We empirically demonstrate the effectiveness of our method on the Spider cross-domain semantic parsing benchmark.

\subsection{Problem Setup}
In cross-domain Text-to-SQL semantic parsing, we want to parse a natural language question $q$ into its corresponding SQL query $r$, given the underlying domain $D$ (the database).

We are given a set of domains 
$\mathcal{D}_{train}=\{D_i=(\mathcal{A}_d, {G}_d)\}_{d=1}^{N_{train}}$, where each domain has a set of annotated pairs of natural language questions and SQL queries $\mathcal{A}_d=\{(q^d_i, r^d_i)\}_{i=1}^{\lvert \mathcal{A}_d \rvert}$, and a schema ${G}_d$.

The schema consists of entities $\mathcal{E}_d=\{e^d_k\}_{k=1}^{\lvert \mathcal{E}_d \rvert}$ belonging to the domain (represented by ``table.column" for example), as well as relationships (links) $\mathcal{L}_d = \{(e^d_k, e^d_l, t)\}$, where $t$ denotes the type of relationship, such as a foreign key or primary key. And each query $r^d_i$ contains one or more entities denoted by $\pmb{e}^d_i$ which is a sequence of entities $e^d_k$'s concatenated in the order they appear in the query $r^d_i$. The corresponding question also implies the same set of entities, although possibly written differently. 

The goal is to train on $\mathcal{D}_{train}$ and generalize to questions on unseen domains $\mathcal{D}_{test}$ given only their schema. 

We first train a baseline semantic parser $P_{\theta}^{\text{teacher}}$ on the training set via maximum likelihood, following the setup of \citet{xu2021optimizing}. Then we look to improve it further using data augmentation.

\subsection{Method Details}

\begin{algorithm}[t]
\caption{{\sc H-NeurSyn}}
\begin{algorithmic}[1]
    \Require training domains $\mathcal{D}_{train}$ and sample sizes $s_1$, $s_2$.
    \LineComment{Phase-I: traing synthesizer components}
    \State Train entity sampler $P_{\phi}(\pmb{e}|G)$ on $\mathcal{D}_{train}$
    \State Finetune a T5 as question generator $P_{\omega}(q|\pmb{e})$ on entities-question pairs extracted from $\mathcal{D}_{train}$
    \State Train a base parser $P_{\theta}(r|q, G)$ on $\mathcal{D}_{train}$
    \LineComment{Phase-II: synthesize}
    \State $D_{aug} \leftarrow \varnothing$
    \For {$G_d \in D_{train}$}
        \State $\{\pmb{e}_j\}_{j=1}^{s_1} \sim p_\phi(\pmb{e}|G_d)$

        \For {$\pmb{e}_j \in \{\pmb{e}_j\}_j$}
            \State $\{q_l\}_l \!\leftarrow\! \texttt{BEAM-SEARCH}(P_{\omega}(.|\pmb{e}_j))$  
            \State $\{r_l\}_l\!\leftarrow\! \texttt{PRED}(P_{\theta},\!\{q_l\}_l, G_d)$
            \State $\{(q_l,\!r_l)\}_{l=1}^{s_2} \!\leftarrow\! \texttt{DEDUP}(\{(q_l,\! r_l)\}_l, \mathcal{A}_d)$
            \State $D_{aug} \leftarrow D_{aug} \cup \{(q_l, r_l)\}_{l=1}^{s_2})$

        \EndFor
   \EndFor
\State $\texttt{return} \;\; D_{aug}$
\end{algorithmic}
\label{alg:highlevel}
\end{algorithm}

As described earlier, we hierarchically decompose the synthesis, starting with sampling entities given the domain first. 
Overall, for a given domain described by its schema $G$, to synthesize data, our method can be described as the following auto-regressive factorization:
$P(q, r, \pmb{e} | G) = P(\pmb{e}|G) P(q|\pmb{e}, G) P(r|q, \pmb{e}, G)$, from which we can sample $(q, r, \pmb{e})$ and keep only the $(q, r)$-pairs for data augmentation. For brevity, we dropped the domain index $d$ henceforth.

Furthermore, we make two conditional independence assumptions to simplify the factorization: $P(q|\pmb{e}, G) = P(q|\pmb{e})$ and $P(r|q, \pmb{e}, G)=P(r|q, G)$. 

The first independence is a good approximation because we can include the domain name in the entity sequence to inform the context of the question. Together with the sequence of table and column names being sampled for the question $q$, there is little additional information in the schema $G$ that is relevant to the $q$. This allows us to fine-tune a pre-trained seq2seq model $P_{\omega}$ for $P(q|\pmb{e})$. 

The second assumption is reasonable as the entity information is expected to be captured explicitly in $q$ or implicitly via the interplay of $q$ and $G$ together. This allows us to use the baseline teacher parser $P_{\theta}$ as a proxy for $P(r|q, G)$. Hence, overall our hierarchical synthesis model is $P(q, r, \pmb{e} | G) \approx P_{\phi, \omega, \theta} (q, r, \pmb{e} | G) = P_{\phi}(\pmb{e}|G) P_{\omega}(q|\pmb{e}) P_{\theta}(r|q, G)$.

Our synthesis procedure {\sc H-NeurSyn} is detailed in Algorithm~\ref{alg:highlevel}. The process starts by first training each of $P_{\phi}(\pmb{e}|G), P_{\omega}(q|\pmb{e}), P_{\theta}(r|q, G)$. Then generate in order $G \rightarrow \pmb{e} \rightarrow q \rightarrow r$. 
For each entity set, we perform beam search to get a large number of questions (line $9$). For each of those questions, we keep only the top-$1$ scoring prediction from beam-search that can successfully execute ($\texttt{PRED}$ line $10$).
Afterward, the examples are deduplicated ($\texttt{DEDUB}$) both within the sampled set as well as with respect to the training data. Finally, for any questions $q$'s with the same logical form $r$, we keep only one question, i.e.\ removing paraphrases via $\texttt{NO-PARA}$.

\begin{table*}[t]
    \centering
    \small
    \begin{tabular}{p{0.52\linewidth} p{0.4\linewidth}}
    \toprule
   \textbf{Input} & \textbf{Output} \\ 
    \toprule
   \texttt{department management : head name text | head age number | head born state text}
     & List the name, born state and age of the heads of departments ordered by age.
    \\ \midrule
    \texttt{culture company : movie year number | movie director text}
    & 
    Which directors had a movie in either 1999 or 2000?
    \\ \bottomrule
    \end{tabular}
    \caption{Input and Output of the Generator}
    \label{tab:ent_q}
\end{table*}

\paragraph{Entity Sampler}

The entity sampler is an autoregressive model given the domain schema: 
\begin{equation}
    p_\phi(\pmb{e}|G) = \prod_{t}p_\phi(e_t| e_{<t}, G) 
\end{equation}%
where $\pmb{e} = [\text{<DB_NAME>},\!e_1\!\ldots\!e_t\!\ldots\!,\text{<EOS>}]$, with $e_1\!\ldots\!e_t\!\ldots$ denoteing the sequence of entities as they appear in the SQL query; prefixed by the domain/database name $e_0=\text{<DB_NAME>}$, and with termination marked by the special $\text{<EOS>}$ token. We model the entities as a sequence rather than set because the order they appear in the query sometimes carries some information about the input question. Given schema $G$, the domain name is deterministically given, so $P(e_0|G)$ is $1$ if $e_0$ is the databse name corresponding to $G$, and $0$ otherwise.

The model is trained via standard maximum likelihood with teacher-forcing. The training examples are entity sequences extracted from ground-truth training queries on each domain, with repeating entities in the same query deduplicated.  

As the model is conditioned on the domain schema $G$, and the parameters $\phi$ are re-used across different domains, 
the sampler can learn to sample entities on both seen and unseen domains.

\paragraph{Question Generator}
For the question generator, we fine-tune a T5 on a small dataset consisting of entity sequences and questions from the training set. 
We use the human-friendly version for the entity names (with underscores removed). See Table \ref{tab:ent_q} for examples of input and outputs.

\subsection{Training with augmented data}

To combine training on the augmented data and the original training set, we initialize a new parser (the student) and 
train it with the following multi-task loss with 
hyper-parameter $\alpha$:
$\mathcal{L}_{\text{MLE}}(\theta) = \mathcal{L}_{\text{MLE}}^{train}(\theta)  + \alpha \times \mathcal{L}_{\text{MLE}}^{aug}(\theta)$.  
Additionally, for each augmented sample, we prepend a special $[\text{AUG}]$ token, inspired by tagged back-translation \cite{caswell2019tagged}.

\paragraph{Zero-shot augmentation}
Furthermore, because all component models of {\sc H-NeurSyn} can generalize across domains, the overall data augmentation can also be applied in the zero-shot setting: given an unseen domain with only its schema $G_{\text{new}}$ available, not the questions, we can synthesis new examples by following the {\sc H-NeurSyn} pipeline.
In this transductive case, 
we can optionally modify the 
loss function by weighting the test data term differently with $\alpha_{new}$ from the training data term with $\alpha_{train}$.

\section{Experiments}

\subsection{Data set and evaluation}
To evaluate {\sc H-NeurSyn} we apply it to the popular and challenging
Spider cross-domain text-to-SQL benchmark data set \cite{yu2018spider}.
The data set contains $10,181$ questions and $5,693$ queries covering 200 databases in 138 domains. As with many previous works \cite{shi2020learning,zhao2021gp}, on the test set, we just consider the Exact Match (EM) evaluation, which only requires models to predict the logical form with table and columns correctly, but not the values. On the dev set, in addition to Exact Match (EM), we can also evaluate Execution Accuracy (EX) by using the ground-truth values. EX is less noisy as it accepts equivalent queries that execute to the same results. 

\subsection{Implementation details}

For both the teacher and student semantic parser models, we follow the model and hyperparameter setup in \citet{xu2021optimizing}. 
The model consists of a deep transformer plus relation-aware transformer module that encodes the question and schema jointly. Our backbone semantic parser model \citet{xu2021optimizing} is comparable to the state-of-the-art method at the time of writing 
listed in the Spider leaderboard. 

The entity sampler is also an encoder-decoder neural net, where the encoder for schema representation is the same as the one used in the semantic parser.
For the decoder, we use a single layer LSTM with pointer net output over the available columns of the domain, i.e.\ the column prediction part of the semantic parser used in \citet{xu2021optimizing, xu2021turing}. The sampler is trained from scratch, with the same training hyperparameter as the parser.

We fine-tune a T5-base \cite{raffel2020exploring} as the question generator without warm-up steps with 
a learning rate $3e-4$, a batch size of $8$ for $3$ epochs, and gradient clipping with a threshold $1.0$. 

For comparisons, we benchmark GAZP \cite{zhong-etal-2020-grounded}  and L2S \cite{wang-etal-2021-learning-synthesize}, both in their original setup, as well as using their augmented data with our (stronger) baseline semantic parser for a fairer comparison. For L2S, we use the augmented data set publicly released by the authors.
For GAZP, we use the publicly released code\footnote{https://github.com/vzhong/gazp} to generate the augmented datasets. For the cycle consistency check required
by GAZP, we implement our own DT-Fixup-based parser. We apply our deduplication filtering for both GAZP and L2S too. 

\subsection{Main Results}
\paragraph{SOTA and Near-SOTA performance}
From Table \ref{tab:spider}, the first important observation is that our method, {\sc H-NeurSyn}, achieves state-of-the-art dev accuracy and near state-of-the-art test accuracy. Because the Spider test set has only $2147$ examples, the gap of $0.1\%$ and $.2\%$ to \citet{scholak2021picard} and \citet{cao2021lgesql} only represent $2$ and $4$ extra mistakes by our method.

\begin{table}[t]
\centering
\small
\begin{tabular}{l l l l l}
\hline

 & \multicolumn{2}{c}{\textbf{Dev Set}} & \multicolumn{2}{c}{\textbf{Test Set}} \\
\bf Model & \bf EM \bf  & \bf EX & \bf EM   \\ \hline
\citet{wang2019rat} & $69.7$   & - & $65.6$  \\
\citet{yu2020grappa} & $73.4$  &- & $69.6$ \\
\citet{shi2020learning} & $71.8$  &- & $69.7$  \\
\citet{zhao2021gp} & $72.8$  & -& $69.8$ \\
\citet{cao2021lgesql}  & $75.1$  & -& $\mathbf{72.0}$ \\ 
\citet{scholak2021picard} & $75.5$ & -& $71.9$  \\ 
\hdashline
L2S & $71.9$   & $72.5$&-   \\
GAZP & $59.1$   & $59.2$  &  $53.3$  \\
\hline
DT-Fixup  & $75.0$ & $74.6$ & ${70.9}$ \\ 
 + L2S(train)  & $75.1$& $73.5$&-  \\
 + GAZP(dev)  & $74.8$& $74.0$ &- \\
 + GAZP(train+dev)  & $76.0$& $74.9$ &-  \\
  + Ours(train)  & $76.4$& - &- \\
 + Ours(train+dev) & $\mathbf{77.2}$ & $\mathbf{76.1}$ & $71.8$\\\hline
\end{tabular}
\caption{Accuracy on the Spider development and test sets, as compared to the other approaches at the top of the Spider leaderboard as of November $28$th, 2021. }  
\label{tab:spider}
\end{table}

\begin{table}[t!]
\centering
% \scriptsize
\small
\begin{tabular}{l c c c c c}
\hline
\bf \# Train & $1694$ &  $2777$ & $1461$ & $1068$ & $7000$\\
\bf \# Test & $248$ &  $446$ & $174$ & $166$ & $1034$\\ \hline
 & \bf Easy & \bf Medium & \bf Hard & \bf Extra & \bf All \\ \hline
\bf DT-Fixup & $91.9$ & $80.9$ & $60.3$ & $48.8$ & $75.0$ \\ 
% \hdashline
\bf +Ours & $\mathbf{92.7}$ & $\mathbf{82.3}$ & $\mathbf{65.5}$ & $\mathbf{52.4}$ & $\mathbf{77.2}$ \\\hline
\hline
\end{tabular}
\caption{Accuracy on Spider by hardness levels.}
\label{tab:breakdown}
\end{table}

On the developement set, without zero-shot augmentation, we achieve $76.4$ (``Ours(train)"), while $77.2$ with it (``Ours(train+dev)"). Note that, our number on the test set could be even higher if we apply the zero-shot version because the test domain schema is available to the inference model. In practice, due to the compute-time limitation of the evaluation script, we cannot apply zero-shot augmentation. However, we believe it is relevant for bootstrapping semantic parsers in real-world applications, so the improvements on the developement set holds practical significance. 

\paragraph{Improvement from data augmentation}
Although the GAZP, L2S each produced improvements over their baseline parsers, with our stronger parser, they do not yield significant improvements. The largest is from GAZP applied in the zero-shot setting with augmentation from both training and dev sets, leading to an improvement of $1.0$ (EM) and $0.3$ (EX). On the other hand, using the same number of augmented points as GAZP, {\sc H-NeurSyn} produced an improvement of $2.2$ (EM) and $1.5$ (EX).

Furthermore, as shown by Table \ref{tab:breakdown}, our augmentation method leads to gains across all difficulty levels, where the levels are defined by \citet{yu2018spider}, with more boost in the hard and extra hard categories ($~6$-points on average).  
This is encouraging, as training data is particularly limited in those hard categories, while manual curation or simple heuristics are not feasible to generate more difficult examples. 

    \begin{table*}[ht]
        \small
        \centering
        \begin{tabular}{p{0.16\linewidth} p{0.25\linewidth} p{0.5\linewidth}}
        \toprule
        \textbf{Sampled entities} & \textbf{Generated Question} & \textbf{Self-labeld SQL} \\ 
        \toprule
        \texttt{perpetrator. location}
        & What are the different locations for perpetrators?
         & 
        \texttt{SELECT DISTINCT perpetrator.location FROM perpetrator}\\
       \cline{2-3}
       & List the locations of perpetrators in ascending alphabetical order.
       &
       \texttt{SELECT perpetrator.location FROM perpetrator ORDER BY perpetrator.location}\\ 
        \cline{2-3}
    
        \midrule
        \texttt{captain.name, captain.age}
        & What are the names and ages of all captains?
        & 
        \texttt{SELECT captain.name, captain.age FROM captain}\\
        \cline{2-3}
       & What is the name of the youngest captain?
       &
       \texttt{SELECT captain.name FROM captain ORDER BY captain.age}\\ 
       \bottomrule
        \end{tabular}
        \caption{Effect of beam search in the question generator}
        \label{tab:beam}
    \end{table*}

\subsection{Analysis of synthesized data}
\label{sec:sample}

To gain insights about why {\sc H-NeurSyn}'s augmented data improves semantic parsing, let us inspect some statistics of the SQL sketches and the columns. 

First, we convert the SQL queries to  "sketches" by masking off the table and column names (values are already not being considered in the target SQL queries). Second, we use a handcrafted rule system to split complex queries into simpler parts, like the template system of \cite{elgohary2020speak} or the shallow grammar of \cite{xu2021turing}. 

Then treating the deconstructed SQL sketches as discrete categories, we can compute the normalized entropy of sketches for both the original and augmented datasets. Similarly, we can compute the same statistic for the table sets and columns sets occurring in the queries. The normalized entropy $\tilde{H}_{P}$ of distribution $P$ is the ratio of its entropy over that of a uniform distribution $U$ over the same space, i.e.\ $\tilde{H}_{P} = H(P)/H(U)$. 

Furthermore, we can evaluate the normalized mutual information of SQL sketches with the table or column entities, where the normalized mutual information is defined as
$\tilde{I}_{X:Y}= {2 I(X:Y) } / {(H(X) + H(Y))}$.

The normalized entropy gives a measure of diversity, whereas the normalized mutual information reflects statistical association. Both are standardized so that they can be averaged across the different databases on Spider, which is shown in Table~\ref{tab:stats}. 
We see that the augmentation improves the diversity of columns $\tilde{H}_{Col}$ by a large margin, while slightly reducing the diversity of sketches $\tilde{H}_{Sketch}$. More importantly, the normalized mutual information between sketch and entities are reduced ($\tilde{I}_{DB:Sketch}$ and $\tilde{I}_{Col:Sketch}$), suggesting that the augmented data potentially can help break some of the spurious correlation that occurs in the original dataset between the logical form and entity occurrences. 

\begin{table}[t]
\centering
%  \scriptsize
\small
\begin{tabular}{l c c | c c}
\hline
\bf Model & \bf Train & \bf Train$_{Aug}$ & \bf Dev & \bf Dev$_{Aug}$\\\hline
\# instances & 7000 &  \textbf{44785} & 1007 & \textbf{6736}\\
%\# unique DBs & 140 &  140 & 20 & 20\\
\# unique sketch & 267 & \textbf{343} & 106 & \textbf{166} \\ 
\# unique col set & 1251 & \textbf{11398} & 181 & \textbf{1748} \\ \hline
$\tilde{H}_{DB} \uparrow$ & 0.961 &\textbf{0.996} & 0.942 & \textbf{0.995} \\ 
$\tilde{H}_{Col} \uparrow$ & 0.483 & \textbf{0.962} & 0.512 & \textbf{0.958} \\ 
$\tilde{H}_{Sketch} \uparrow$ & \textbf{0.747} & 0.644 & \textbf{0.820} & 0.702  \\ 
$\tilde{I}_{DB:Sketch} \downarrow$ & {0.290} & \textbf{0.108} & {0.307} & \textbf{0.108} \\ 
$\tilde{I}_{Col:Sketch} \downarrow$ & {0.510} & \textbf{0.396} & {0.531} & \textbf{0.438}  \\ 
\hline
\end{tabular}
\caption{Statistics of datasets in Spider and Spider-aug.}
\label{tab:stats}
\end{table}

Qualitatively, Table \ref{tab:beam} shows some samples from the question generator. Given a fixed entity set, there are various ways to generate questions with different logical forms. We find that in most cases, rather than paraphrasing, the generator learns to 
effectively sweep through the possible logical forms given the entities, which
leads to a reduction of the correlation between entities and SQL sketches, as shown by $\tilde{I}_{DB:Sketch}$ and $\tilde{I}_{Col:Sketch}$ of Table~\ref{tab:stats}.

\begin{table}[t]
\centering
\small
\begin{tabular}{p{0.65\linewidth} p{0.2\linewidth} }
\hline
\bf Model & \bf Dev EM\\ \hline
Backbone (DT-Fixup w/ small encoder)  & $72.7 \pm 0.4 $\\
\hline
 \makecell[l]{\textbf{Ours} \\ 1. learnable(train+dev) \\ 2. $s_1$=80, $s_2$=20 \\ 3. $\alpha_{train}$=0.3, $\alpha_{dev}$=0.1 \\4. w/ \texttt{DEDUP} filtering} & \bm{$75.2 \pm 0.3$} \\\hline
\hline
\it  Augmentation Baselines \\ 
L2S(train) w/o \texttt{DEDUP} filtering & $72.3 \pm 0.2$  \\
GAZP(dev) w/o \texttt{DEDUP} filtering & $73.1 \pm 0.4$  \\
L2S(train) & $72.5 \pm 0.3$  \\
GAZP(dev) & $73.2 \pm 0.5$  \\
GAZP(train+dev) & $74.2 \pm 0.1$\\ \hline
\it Sampling Strategies \\ 
Ours-random(train) & $73.0 \pm 0.2$\\
Ours-learnable(train) & $75.0 \pm 0.3$  \\
Ours-learnable(dev) & $74.4 \pm 0.3$\\\hline
\it Question Diversity  \\ 

$s_1$=80, $s_2$=1 & $74.0 \pm 0.5$\\
$s_1$=40, $s_2$=20 & $75.0 \pm 0.4$\\
$s_1$=80, $s_2$=40 & $75.1 \pm 0.2$\\

\hline
\it Joint Training \\ 

$\alpha_{train}$=0.3, $\alpha_{dev}$=0.3 & $74.7 \pm 0.1$\\
$\alpha_{train}$=0.1, $\alpha_{dev}$=0.1 & $74.6 \pm 0.0$\\
Pretrain-Finetune & $73.6\pm 0.1$\\
Combine & $73.6\pm 0.4$\\
\hline
\it Filtering \\ 

w/o \texttt{DEDUP} filtering & $73.7 \pm 0.6$  \\
random down-sampling & $74.3 \pm 0.3$  \\
\hline

\end{tabular}
\caption{Alation Study}
\label{tab:ablation}

\end{table}

\subsection{Ablation study}

We analyze to demonstrate the importance of each aspect of our approach in Table \ref{tab:ablation}.
In all experiments, we use the smaller encoder due to computing limitations: RoBERTa-base with $8$ relation-aware transformer (RAT) layers \cite{wang2019rat}, instead of RoBERTa-large with $24$ RAT layers. 

On sampling strategy, we compare to uniform-randomly sampling 
entities from the schema, instead of the learned sampler. 
Unsurprisingly, the learned sampler is superior to random sampling, as it captures a more natural combination of entities. Table \ref{tab:random-learnable} gives more insights about why learned sampling is crucial: it is often impossible to formulate a natural sentence that makes sense from randomly sampled entities, as shown in Table \ref{tab:random-learnable}.

The hyper-parameters $s_1$ and $s_2$ control the diversity of entities $E$ and logics $L$. From Table \ref{tab:ablation} we can see tuning them at a reasonable range does not affect the augmentation performance much. However, the performance drops dramatically when $s_2$=1 (i.e. no beam search for question generation). 

Tuning the student training hyper-parameters $\alpha_{train}$ and $\alpha_{dev}$ moderately affects the performance, while the best performance is achieved when  $\alpha_{train}$=0.3 and  $\alpha_{dev}$=0.1. This is because the quality of augmented examples on the dev set databases is relatively lower than those on the training set databases due to sampling error and self-labeling error, as neither sampler nor parser is trained on the dev set domains. Besides, we also compare the training strategies (Pretrain-Finetune and Combine) employed by GAZP and L2S, and the results show neither of them works for our data augmentation method. Note that both the main results in Table \ref{tab:spider} and ``Augmentation Baselines" in Table \ref{tab:ablation} confirm that our augmentation method itself is superior to GAZP and L2S even after controlling for other aspects of improvements.

\begin{table}[t]
    \small
    \centering
    \begin{tabular}{p{0.1\linewidth} p{0.43\linewidth} p{0.36\linewidth}}
    \toprule
    \textbf{} & \textbf{Sampled entities} & \textbf{Generated questions} \\ 
    \toprule
    Random
    & \texttt{people.name, church.church_id, wedding.male_id,  church.organized_by}
     & Show the names of people and the organized by the church they belong to .
    \\ \midrule
    Learned
    & \texttt{people.people_id, people.name, people.is_male, wedding.male_id}
    & 
    What are the names of people who are male and have never attended a wedding ?
    \\ \midrule
    Learned
    & \texttt{church.church_id, church.name, wedding.church_id, wedding.year}
    & 
    What is the name of the church that hosted the largest number of weddings in year 2010 ?
    \\ \bottomrule
    \end{tabular}
    \caption{Random v.s. learned entity sampler}
    \label{tab:random-learnable}
\end{table}

\section{Conclusion}
 In this work, we propose a data augmentation method for the cross-domain text-to-SQL task. Being a fully neural method, our approach avoids burdensome grammar engineering, while yielding more performance gain than previous data augmentation methods. While designed for semantic parsing, we believe our hierarchical synthesis approach coupled with a teacher-student framework could be applied more broadly in other problems such as machine reading comprehension, question answering over knowledge base, and visual question answering.

\clearpage

\bibliography{custom}
\bibliographystyle{acl_natbib}

\end{document}